\documentclass[11pt,a4paper]{article}
\usepackage[hyperref]{conll-2019}
\usepackage{cite}
\usepackage{amsmath,amssymb,amsfonts}
\usepackage{algorithmic}
\usepackage{graphicx}
\usepackage{textcomp}
\usepackage{xcolor}

\usepackage{latexsym}
\usepackage{xspace}
\usepackage{url}
\usepackage{booktabs} 
\usepackage{siunitx}  
\usepackage{todonotes}
\usepackage[font={bf}, tableposition=top]{caption}     
\usepackage{multirow}

\newcommand{\tsl}{T1\xspace}
\newcommand{\tds}{T2\xspace}
\newcommand{\concat}{\ensuremath{\mspace{2.5mu}\Vert\mspace{2.5mu}}\xspace}
\newcommand{\ensemble}{\ensuremath{\oplus}\xspace}

\newenvironment {squishlist}
{\begin{list}{$\bullet$}
  { \setlength{\itemsep}{0pt}
     \setlength{\parsep}{3pt}
     \setlength{\topsep}{3pt}
     \setlength{\partopsep}{0pt}
     \setlength{\leftmargin}{1.5em}
     \setlength{\labelwidth}{1em}
     \setlength{\labelsep}{0.5em} } }
{\end{list}}

\usepackage[hyperref]{conll-2019}
\usepackage{times}
\usepackage{latexsym}

\usepackage{url}

\aclfinalcopy 

\title{Predicting the Role of Political Trolls in Social Media}

\author{Atanas Atanasov \\
  Sofia University  \\
  Bulgaria\\
  {\tt amitkov@uni-sofia.bg} \\\And
  Gianmarco De Francisci Morales \\
  ISI Foundation \\
  Italy \\
  {\tt gdfm@acm.org} \\\And
  Preslav Nakov \\
  Qatar Computing Research\\
  Institute, HBKU, Qatar \\
  {\tt pnakov@qf.org.qa} \\}

\date{}

\begin{document}
\maketitle
\begin{abstract}
We investigate the political roles of ``Internet trolls'' in social media. 
Political trolls, such as the ones linked to the Russian Internet Research Agency (IRA), have recently gained enormous attention for their ability to sway public opinion and even influence elections.
Analysis of the online traces of trolls has shown different behavioral patterns, which target different slices of the population.
However, this analysis is manual and labor-intensive, thus making it impractical as a first-response tool for newly-discovered troll farms.
In this paper, we show how to automate this analysis by using machine learning in a realistic setting.
In particular, we show how to classify trolls according to their political role ---left, news feed, right--- by using features extracted from social media, i.e., Twitter, in two scenarios:
(\emph{i})~in a traditional supervised learning scenario, where labels for trolls are available, and (\emph{ii})~in a distant supervision scenario, where labels for trolls are not available, and we rely on more-commonly-available labels for news outlets mentioned by the trolls.
Technically, we leverage the community structure and the text of the messages in the online social network of trolls represented as a graph, from which we extract several types of learned representations, i.e.,~embeddings, for the trolls.
Experiments on the ``IRA Russian Troll'' dataset show that our methodology improves over the state-of-the-art in the first scenario, while providing a compelling case for the second scenario, which has not been explored in the literature thus far.
\end{abstract}

\section{Introduction}

Internet ``trolls'' are users of an online community who quarrel and upset people, seeking to sow discord by posting inflammatory content.
More recently, organized ``troll farms'' of political opinion manipulation trolls have also emerged.

\noindent Such farms usually consist of state-sponsored agents who control a set of pseudonymous user accounts and personas, the so-called ``sockpuppets'', which disseminate misinformation and propaganda in order to sway opinions, destabilize the society, and even influence elections \citep{IRA:2018}.

The behavior of political trolls has been analyzed in different recent circumstances, such as the 2016 US Presidential Elections and the Brexit referendum in UK \citep{IRA:2018,llewellyn2018whom}.
However, this kind of analysis requires painstaking and time-consuming manual labor to sift through the data and to categorize the trolls according to their actions.
Our goal in the current paper is to automate this process with the help of machine learning (ML).
In particular, we focus on the case of the 2016 US Presidential Elections, for which a public dataset from Twitter is available.
For this case, we consider only accounts that post content in English, and we wish to divide the trolls into some of the functional categories identified by \citet{IRA:2018}: \emph{left troll}, \emph{right troll}, and \emph{news feed}. 

We consider two possible scenarios.
The first, prototypical ML scenario is supervised learning,
where we want to learn a function from users to categories \{\emph{left}, \emph{right}, \emph{news feed}\},
and the ground truth labels for the troll users are available.
This scenario has been considered previously in the literature by \citet{kim2019tracking}.
Unfortunately, a solution for such a scenario is not directly applicable to a real-world use case.
Suppose a new troll farm trying to sway the upcoming European or US elections has just been discovered.
While the identities of the accounts might be available, the labels to learn from would not be present. Thus, any supervised machine learning approach would fall short of being a fully automated solution to our initial problem.

\noindent A more realistic scenario assumes that labels for troll accounts are \emph{not available}.
In this case, we need to use some external information in order to learn a labeling function.
Indeed, we leverage more persistent entities and their labels: news media.
We assume a learning scenario with distant supervision where labels for news media are available.
By combining these labels with a citation graph from the troll accounts to news media,
we can infer the final labeling on the accounts themselves without any need for manual labeling.

One advantage of using distant supervision is that we can get insights about the behavior of a newly-discovered troll farm quickly and effortlessly.
Differently from troll accounts in social media, which usually have a high churn rate, news media accounts in social media are quite stable.
Therefore, the latter can be used as an anchor point to understand the behavior of trolls, for which data may not be available.

We rely on embeddings extracted from social media.
In particular, we use a combination of embeddings built on the user-to-user mention graph, the user-to-hashtag mention graph, and the text of the tweets of the troll accounts.
We further explore several possible approaches using label propagation for the distant supervision scenario.

As a result of our approach, we improve the classification accuracy by more than 5 percentage points for the supervised learning scenario.
The distant supervision scenario has not previously been considered in the literature, and is one of the main contributions of the paper.
We show that even by hiding the labels from the ML algorithm, we can recover 78.5\% of the correct labels.

The contributions of this paper can be summarized as follows:
\begin{squishlist}
    \item We predict the political role of Internet trolls (\emph{left}, \emph{news feed}, \emph{right}) in a realistic, unsupervised scenario, where labels for the trolls are not available, and which has not been explored in the literature before.
    \item We propose a novel distant supervision approach for this scenario, based on graph embeddings, BERT, and label propagation, which projects the more-commonly-available labels for news media onto the trolls who cited these media. 
    \item We improve over the state of the art in the traditional, fully supervised setting, where training labels are available.
\end{squishlist}

\section{Related Work}
\label{sec:related}

\subsection{Trolls and Opinion Manipulation}
The promise of social media to democratize content creation \citep{kaplan2010users}
has been accompanied by many malicious attempts to spread misleading information over this new medium, which quickly got populated by \emph{sockpuppets} \citep{Kumar:2017:AMS:3038912.3052677}, \emph{Internet water army} \citep{Chen:2013:BIW:2492517.2492637}, \emph{astroturfers} \citep{Ratkiewicz:2011:TMS:1963192.1963301}, and \emph{seminar users} \citep{SeminarUsers2017}. 
Several studies have shown that trust is an important factor in online relationships \citep{DBLP:journals/intr/HoKL12,
DBLP:journals/intr/Ku12,
DBLP:journals/intr/HsuCH14,
DBLP:journals/intr/ElbeltagiA16,
DBLP:journals/intr/HaJJC16}, but building trust is a long-term process and our understanding of it is still in its infancy \citep{salo2007conceptual}. 
This makes it easy for politicians and companies to manipulate user opinions in forums \citep{Dellarocas06,DBLP:journals/intr/LiGWZ16,ZHUANG201824}.

\emph{Trolls.} Social media have seen the proliferation of fake news and clickbait \citep{Hardalov2016,RANLP2017:clickbait}, aggressiveness \citep{moore2012anonymity}, and trolling \citep{cole2015s}. The latter often is understood to concern malicious online behavior that is intended to disrupt interactions, to aggravate interacting partners, and to lure them into fruitless argumentation in order to disrupt online interactions and communication \citep{Chen:2013:BIW:2492517.2492637}.
Here we are interested in studying not just any trolls, but those that engage in opinion manipulation \citep{Mihaylov2015FindingOM,Mihaylov2015ExposingPO,InternetResearchJournal:2018}.
This latter definition of \emph{troll} has also become prominent in the general public discourse recently.
\citet{del2016spreading} have also suggested that the spreading of misinformation online is fostered by the presence of polarization and echo chambers in social media
\citep{garimella2016quantifying,garimella2017effect,garimella2018political}.

\emph{Trolling behavior} is present and has been studied in all kinds of online media: online magazines \citep{binns2012don}, social networking sites \citep{cole2015s}, online computer games \citep{thacker2012exploratory}, online encyclopedia \citep{shachaf2010beyond}, and online newspapers \citep{ruiz2011public}, among others.

\emph{Troll detection} was addressed using 
domain-adapted sentiment analysis \citep{Seah2015}, lexico-syntactic features about writing style and structure 
\citep{chen2012detecting,mihaylov-nakov:2016:P16-2}, and graph-based approaches over signed social networks
\citep{kumar2014accurately}. 

\emph{Sockpuppet} is a related notion, and refers to a person who assumes a false identity in an Internet community and then speaks to or about themselves while pretending to be another person.
The term has also been used to refer to opinion manipulation, e.g., in Wikipedia \citep{SolorioHM14}.
Sockpuppets have been identified by using authorship-identification techniques and link analysis \citep{Bu:2013:SPD:2400768.2401510}.
It has been also shown that sockpuppets differ from ordinary users in their posting behavior, linguistic traits, and social network structure \citep{Kumar:2017:AMS:3038912.3052677}.

\emph{Internet Water Army} is a literal translation of the Chinese term \emph{wangluo shuijun}, which is a metaphor for a large number of people who are well organized to flood the Internet with purposeful comments and articles.
Internet water army has been allegedly used in China by the government (also known as \emph{50 Cent Party}) as well as by a number of private organizations. 

\emph{Astroturfing} is an effort to simulate a political grass-roots movement.
It has attracted strong interest from political science, and research on it has focused on massive streams of microblogging data
\citep{Ratkiewicz:2011:TMS:1963192.1963301}.

\emph{Identification of malicious accounts} in social media includes detecting spam accounts \citep{almaatouq2016if,mccord2011spam}, fake accounts \citep{fire2014friend,cresci2015fame}, compromised and phishing accounts \citep{adewole2017malicious}. Fake profile detection has also been studied in the context of cyber-bullying \citep{galan2016supervised}.
A related problem is that of \emph{Web spam detection}, which has been addressed as a text classification problem \citep{sebastiani2002machine}, e.g., using spam keyword spotting \citep{dave2003mining}, lexical affinity of arbitrary words to spam content \citep{hu2004mining}, frequency of punctuation and word co-occurrence \citep{li2006combining}. 

\emph{Trustworthiness} of online statements is an emerging topic, given the interest in fake news \citep{Lazer1094}.
It is related to trolls, as they often engage in opinion manipulation and spread rumors  \citep{Vosoughi1146}.
Research topics include predicting the credibility of information in social media \citep{ma2016detecting,mitra2017parsimonious,RANLP2017:factchecking:external,Popat:2017:TLE:3041021.3055133} and political debates \citep{Hassan:15,RANLP2017:debates, NAACL2018:claimrank},
and stance classification \citep{NAACL2018:stance}.

\noindent For example, \citet{Castillo:2011:ICT:1963405.1963500} leverage user reputation, author writing style, and various time-based features,
\citet{Canini:2011} analyze the interaction of content and social network structure, and \citet{morris2012tweeting} studied how Twitter users judge truthfulness.
\citet{PlosONE:2016} study how people handle rumors in social media, and found that users with higher reputation are more trusted, and thus can spread rumors easily.
\citet{lukasik-cohn-bontcheva:2015:ACL-IJCNLP} use temporal patterns to detect rumors and to predict their frequency, and \citet{PlosONE:2016} focus on conversational threads. More recent work has focused on the credibility and the factuality in community forums \citep{RANLP2017:credibility:trolls,AAAI2018:factchecking,mihaylova-etal-2019-semeval,InternetResearchJournal:2018}.

\subsection{Understanding the Role of Political Trolls}
None of the above work has focused on understanding the role of political trolls. The only closely relevant work is that of
\citet{kim2019tracking}, who predict the roles of the Russian trolls on Twitter by leveraging social theory and Actor-Network Theory approaches.
They characterize trolls 
using the digital traces they leave behind, which is modeled using a time-sensitive semantic edit distance. 

\noindent For this purpose, they use the ``IRA Russian Troll'' dataset \citep{IRA:2018}, which we also use in our experiments. However, we have a very different approach based on graph embeddings, which we show to be superior to their method in the supervised setup.
We further experiment with a new, and arguably more realistic, setup based on distant supervision, where labels are not available. To the best of our knowledge, this setup has not been explored in previous work.

\subsection{Graph Embeddings}
Graph embeddings are machine learning techniques to model and capture key features from a graph automatically.
They can be trained either in a supervised or in an unsupervised manner \citep{cai2018comprehensive}.
The produced embeddings are latent vector representations that map each vertex $V$ in a graph $G$ to a $d$-dimensional vector.
The vectors capture the underlying structure of the graph by putting ``similar'' vertices close together in the vector space.
By expressing our data as a graph structure, we can leverage and extract critical insights about the topology and the contextual relationships between the vertices in the graph.

\noindent In mathematical terms, graph embeddings can be expressed as a function $f: V \rightarrow R^d$ from the set of vertices $V$ to a set of embeddings, where $d$ is the dimensionality of the embeddings.
The function $f$ can be represented as a matrix of dimensions $|V| \times d$.
In our experiments, we train Graph Embeddings in an unsupervised manner by using node2vec \citep{grover2016node2vec}, which is based on random walks over the graph.
Essentially, this is an application of the well-known skip-gram model \citep{Mikolov:2013:DRW:2999792.2999959} from word2vec to random walks on graphs.

Besides \emph{node2vec}, there have been a number of competing proposals for building graph embeddings; see \citep{cai2018comprehensive} for an extensive overview of the topic.
For example, \emph{SNE} \citep{liao2018attributed} model both the graph structure and some node attributes.
Similarly, \emph{Line} \citep{tang2015line} represent each node as the concatenation of two embedded vectors that model first- and second-order proximity.
\emph{TriDNR} \citep{pan2016tri} represents nodes by coupling several neural network models.
For our experiments, we use node2vec, as we do not have access to user attributes: the users have been banned from Twitter, their accounts were suspended, and we only have access to their tweets thanks to the ``IRA Russian Trolls'' dataset.

\section{Method}
\label{sec:method}

Given a set of known political troll users (each user being represented as a collection of their tweets), we aim to detect their role: \emph{left}, \emph{right}, or \emph{news feed}. \citet{IRA:2018} describe these roles as follows:

\textbf{Right Trolls} spread nativist and right-leaning populist messages. Such trolls support the candidacy and Presidency of Donald Trump and denigrate the Democratic Party; moreover, they often send divisive messages about mainstream and moderate Republicans.

\textbf{Left Trolls} send socially liberal messages and discuss gender, sexual, religious, and -especially- racial identity. Many tweets are seemed intentionally divisive, attacking mainstream Democratic politicians, particularly Hillary Clinton, while supporting Bernie Sanders prior to the elections.

\textbf{News Feed Trolls} overwhelmingly present themselves as US local news aggregators, linking to legitimate regional news sources and tweeting about issues of local interest.

\noindent Technically, we leverage the community structure and the text of the messages in the social network of political trolls represented as a graph, from which we learn and extract several types of vector representations, i.e., troll user embeddings. Then, armed with these representations, we tackle the following tasks:
\begin{itemize}
    \item[\textbf{\tsl}] A fully supervised learning task, where we have labeled training data with example troll and their roles;
    \item[\textbf{\tds}] A distant supervision learning task, in which labels for the troll roles are \emph{not} available at training time, and thus we use labels for news media as a proxy, from which we infer labels for the troll users.
\end{itemize}

\subsection{Embeddings}
\label{sec:embeddings}

We use two graph-based (user-to-hashtag and user-to-mentioned-user) and one text-based (BERT) embedding representations.

\subsubsection{U2H}
We build a bipartite, undirected User-to-Hashtag (U2H) graph, where nodes are users and hashtags, and there is an edge $(u,h)$ between a user node $u$ and a hashtag node $h$ if user $u$ uses hashtag $h$ in their tweets.
This graph is bipartite as there are no edges connecting two user nodes or two hashtag nodes.
We run node2vec \citep{grover2016node2vec} on this graph, and we extract the embeddings for the users (we ignore the hashtag embeddings).
We use 128 dimensions for the output embeddings.
These embeddings capture how similar troll users are based on their usage of hashtags.

\subsubsection{U2M}
We build an undirected User-to-Mentioned-User (U2M) graph, where the nodes are users, and there is an edge $(u,v)$ between two nodes if user $u$ mentions user $v$ in their tweets (i.e., $u$ has authored a tweet that contains ``@$v$'' ).
We run node2vec on this graph and we extract the embeddings for the users.
As we are interested only in the troll users, we ignore the embeddings of users who are only mentioned by other trolls.
We use 128 dimensions for the output embeddings.
The embeddings extracted from this graph capture how similar troll users are according to the targets of their discussions on the social network.

\subsubsection{BERT}
BERT offers state-of-the-art text embeddings based on the Transformer \citep{devlin2018bert}.
We use the pre-trained BERT-large, uncased model, which has 24-layers, 1024-hidden, 16-heads, and 340M parameters, which yields output embeddings with 768 dimensions.
Given a tweet, we generate an embedding for it by averaging the representations of the BERT tokens from the penultimate layer of the neural network. To obtain a representation for a user, we average the embeddings of all their tweets.
The embeddings extracted from the text capture how similar users are according to their use of language.

\begin{table*}[t]
\begin{center}
    \small
    \begin{tabular}{lcllp{8.5cm}}
    \toprule
       \textbf{Role} & \bf Users & \textbf{Tweets} & \textbf{User Example} & \textbf{Tweet Example}\\
      \midrule
       Left & 233 & \num{427141} & @samirgooden & @MichaelSkolnik @KatrinaPierson @samesfandiari 
       Trump folks need to stop going on CNN.\\ 
       Right & 630 & \num{711668} & @chirrmorre & BREAKING: 
       Trump ERASES Obama's Islamic Refugee Policy! https://t.co/uPTneTMNM5\\
       News Feed & 54 & \num{598226} & @dailysandiego & Exit poll: Wisconsin GOP voters excited, scared about 
       Trump  \#politics\\
      \bottomrule
    \end{tabular}
    \caption{\label{tab:IRA}Statistics and examples from the IRA Russian Trolls Tweets dataset.}
\end{center}
\end{table*}

\subsection{Fully Supervised Learning (\tsl)}

Given a set of troll users for which we have labels, we use the above embeddings as a representation to train a classifier.
We use an L2-regularized logistic regression (LR) classifier.
Each troll user is an example, and the label for the user is available for training thanks to manual labeling.
We can therefore use cross-validation to evaluate the predictive performance of the model, and thus the predictive power of the features.

We experiment with two ways of combining features: \emph{embedding concatenation} and \emph{model ensembling}.
Embedding concatenation concatenates the feature vectors from different embeddings into a longer feature vector, which we then use to train the LR model.
Model ensembling instead trains a separate model with each kind of embedding, and then merges the prediction of the different models by averaging the posterior probabilities for the different classes.
Henceforth, we denote embedding concatenation with the symbol \concat and model ensembling with \ensemble.
For example, U2H~\concat~U2M is a model trained on the concatenation of U2H and U2M embeddings, while U2H~\ensemble~BERT represents the average predictions of two models, one trained on U2H embeddings and one on BERT.

\subsection{Distant Supervision (\tds)}
In the distant supervision scenario, we assume not to have access to user labels.
Given a set of troll users \underline{without} labels, we use the embeddings described in Section~\ref{sec:embeddings} together with mentions of \emph{news media} by the troll users to create proxy models.
We assume that labels for news media are readily available, as they are stable sources of information that have a low churn rate.

\noindent We propagate labels from the given media to the troll user that mentions them according to the following media-to-user mapping:
\begin{equation}
\begin{array}{r@{}l}
    \label{eq:mapping}
    \emph{LEFT} &\rightarrow \emph{left} \\
    \emph{RIGHT} &\rightarrow \emph{right} \\
    \emph{CENTER} &\rightarrow \emph{news feed}
\end{array}
\end{equation}

This propagation can be done in different ways: (\emph{a})~by training a proxy model for media and then applying it to users, (\emph{b})~by additionally using label propagation (LP) for semi-supervised learning.

Let us describe the proxy model propagation for ($a$) first.
Let $M$ be the set of media, and $U$ be the set of users.
We say a user $u \in U$ mentions a medium $m \in M$ if $u$ posts a tweet that contains a link to the website of $m$.
We denote the set of users that mention the medium $m$ as $C_m \subseteq U$.

We can therefore create a representation for a medium by aggregating the embeddings of the users that mention the target medium.
Such a representation is convenient as it lies in the same space as the user representation.
In particular, given a medium $m \in M$, we compute its representation $R(m)$ as
\begin{equation}
\label{eq:media-repr}
R(m) = \frac{1}{|C_m|} \sum_{u \in C_m} R(u),
\end{equation}
where $R(u)$ is the representation of user $u$, i.e.,~one (or a concatenation) of the embeddings described in Section~\ref{sec:embeddings}.

Finally, we can train a LR model that uses $R(m)$ as features and the label for the medium $l(m)$.
This model can be applied to predict the label of a user $u$ by using the same type of representation $R(u)$, and the label mapping in Equation~\ref{eq:mapping}.

Label Propagation ($b$) is a transductive, graph-based, semi-supervised machine learning algorithm that, given a small set of labeled examples, assigns labels to previously unlabeled examples.
The labels of each example change in relationship to the labels of \emph{neighboring} ones in a properly-defined graph.

More formally, given a partially-labeled dataset of examples $X = X_u \cup X_l$, of which $X_l$ are labeled examples with labels $Y_l$, and $X_u$ are unlabeled examples, and a similarity graph $G(X,E)$, the label propagation algorithm finds the set of unknown labels $Y_u$ such that the number of discordant pairs $(u,v) \in E : y_u \neq y_v$ is minimized, where $y_z$ is the label assigned to example $z$.

The algorithm works as follows:
At every iteration of propagation, each unlabeled node updates its label to the most frequent one among its neighbors.
LP reaches convergence when each node has the same label as the majority of its neighbors.
We define two different versions of LP by creating two different versions of the similarity graph $G$.

\paragraph{LP1} \emph{Label Propagation using direct mention}.\\
In the first case, the set of edges among users $U$ in the similarity graph $G$ consists of the logical OR between the 2-hop closure of the U2H and the U2M graph.
That is, for each two users $u, v \in U$, there is an edge in the similarity graph $(u,v) \in E$ if $u$ and $v$ share a common hashtag or a common user mention
\begin{multline*}
(u,h) \in \text {U2H } \land (v,h) \in \text{ U2H } \lor \\
 (u,w) \in \text{ U2M } \land (v,w) \in \text{ U2M }
\end{multline*}

The graph therefore uses the same information that is available to the embeddings.

To this graph, which currently encompasses only the set of users $U$, we add connections to the set of media $M$.
We add an edge between each pair $(u,m)$ if $u \in C_m$.
Then, we run the label propagation algorithm, which propagates the labels from the labeled nodes $M$ to the unlabeled nodes $U$, thanks to the mapping from Equation~\ref{eq:mapping}.

\paragraph{LP2} \emph{Label Propagation based on a similarity graph}.\\ 
In this case, we use the same representation for the media as in the proxy model case above, as described by Equation~\ref{eq:media-repr}.
Then, we build a similarity graph among media and users based on their embeddings.
For each pair $x,y \in U \cup M$ there is an edge in the similarity graph $(x,y) \in E$ iff
\[
\text{sim}(R(x),R(y)) > \tau, 
\]
where sim is a similarity function between vectors, e.g.,~cosine similarity, and $\tau$ is a user-specified parameter that regulates the sparseness of the similarity graph.

Finally, we perform label propagation on the similarity graph defined by the embedding similarity, with the set of nodes corresponding to $M$ starting with labels, and with the set of nodes corresponding to $U$ starting without labels.

\section{Data}


\subsection{IRA Russian Troll Tweets}
Our main dataset contains \num{2973371} tweets by \num{2848} Twitter users, which the US House Intelligence Committee has linked to the Russian Internet Research Agency (IRA).
The data was collected and published by \citet{IRA:2018}, and then made available online.\footnote{\url{http://github.com/fivethirtyeight/russian-troll-tweets}} 
The time span covers the period from February 2012 to May 2018.

The trolls belong to the following manually assigned roles: Left Troll, Right Troll, News Feed, Commercial, Fearmonger, Hashtag Gamer, Non English, Unknown.
\citet{kim2019tracking} have argued that the first three categories are not only the most frequent, but also the most interesting ones. Moreover, focusing on these troll types allows us to establish a connection between troll types and the political bias of the news media they mention.
Table~\ref{tab:IRA} shows a summary of the troll role distribution, the total number of tweets per role, as well as examples of troll usernames and tweets.

\subsection{Media Bias/Fact Check}
We use data from Media Bias/Fact Check (MBFC)\footnote{\url{http://mediabiasfactcheck.com}} to label news media sites.
MBFC divides news media into the following bias categories: Extreme-Left, Left, Center-Left, Center, Center-Right, Right, and Extreme-Right.
We reduce the granularity to three categories by grouping Extreme-Left and Left as LEFT, Extreme-Right and Right as RIGHT, and Center-Left, Center-Right, and Center as CENTER.

\begin{table}[tbh]
\begin{center}
    \begin{tabular}{lcl}
    \toprule
       \textbf{Bias} & \textbf{Count} & \textbf{Example}\\
      \midrule
       LEFT & 341 & www.cnn.com\\
       RIGHT & 619 & www.foxnews.com\\
       CENTER & 372 & www.apnews.com\\
      \bottomrule
    \end{tabular}
    \caption{\label{tab:MBFC}Summary statistics about the Media Bias/Fact Check (MBFC) dataset.}
\end{center}
\end{table}

Table~\ref{tab:MBFC} shows some basic statistics about the resulting media dataset.
Similarly to the IRA dataset, the distribution is right-heavy.

\begin{table*}[th]
\begin{center}
    \begin{tabular}{l rr rr}
      \toprule
      \multirow{2}{*}{\textbf{Method}} & \multicolumn{2}{c}{\textbf{Full Supervision (\tsl)}} & \multicolumn{2}{c}{\textbf{ Distant Supervision (\tds)}}\\
      \cmidrule(lr){2-3} \cmidrule(lr){4-5}
        & \textbf{Accuracy} & \textbf{Macro F1} & \textbf{Accuracy} & \textbf{Macro F1} \\
        \midrule
        Baseline (majority class) & 68.7 & 27.1 & 68.7 & 27.1\\
        \citet{kim2019tracking} & 84.0 & 75.0 & N/A & N/A\\
        BERT     & 86.9 & 83.1 & 75.1 & 60.5\\
        U2H      & 87.1 & 83.2 & 76.3 & 60.9\\
        U2M      & 88.1 & 83.9 & 77.3 & 62.4\\
        U2H \ensemble U2M & 88.3 & 84.1 & 77.9 & 64.1\\
        U2H \concat   U2M & 88.7 & 84.4 & 78.0 & 64.6\\
        U2H \ensemble U2M \ensemble BERT & 89.0 & 84.4 & 78.0 & 65.0\\
        U2H \concat U2M \concat BERT  & 89.2 & 84.7 & 78.2& 65.1\\
        U2H \concat U2M \concat BERT + LP1 & 89.3 & 84.7 & 78.3 & 65.1\\
        U2H \concat U2M \concat BERT + LP2 & 89.6 & 84.9 & 78.5 & 65.7\\
      \bottomrule
    \end{tabular}
    \caption{\label{tab:tdsvstsl}   Predicting the role of the troll users using full vs. distant supervision.}
\end{center}
\end{table*}

\section{Experiments and Evaluation}
\label{sec:experiment}

\subsection{Experimental Setup}

For each user in the IRA dataset, we extracted all the links in their tweets, we expanded them recursively if they were shortened, we extracted the domain of the link, and we checked whether it could be found in the MBFC dataset.
By grouping these relationships by media, we constructed the sets of users $C_m$ that mention a given medium $m \in M$.

The U2H graph consists of \num{108410} nodes and \num{443121} edges, 
while the U2M graph has \num{591793} nodes and \num{832844} edges.
We ran node2vec on each graph to extract 128-dimensional vectors for each node.
We used these vectors as features for the fully supervised and for the distant-supervision scenarios.
For Label Propagation, we used an empirical threshold for edge materialization $\tau = 0.55$, to obtain a reasonably sparse similarity graph.

We used two evaluation measures: accuracy, and macro-averaged F1 (the harmonic average of precision and recall).
In the supervised scenario, we performed 5-fold cross-validation.
In the distant-supervision scenario, we propagated labels from the media to the users.
Therefore, in the latter case the user labels were only used for evaluation.

\subsection{Evaluation Results}

Table~\ref{tab:tdsvstsl} shows the evaluation results.
Each line of the table represents a different combination of features, models, or techniques.
As mentioned in Section~\ref{sec:method}, the symbol `\concat' denotes a single model trained on the concatenation of the features, while the symbol `\ensemble' denotes an averaging of individual models trained on each feature separately.
The tags `LP1' and `LP2' denote the two label propagation versions, by mention and by similarity, respectively.

\noindent We can see that accuracy and macro-averaged F1 are strongly correlated and yield very consistent rankings for the different models. Thus, henceforth we will focus our discussion on accuracy.

We can see in Table~\ref{tab:tdsvstsl} that it is possible to predict the roles of the troll users by using distant supervision with relatively high accuracy.
Indeed, the results for \tds are lower compared to their \tsl counterparts by only 10 and 20 points absolute in terms of accuracy and F1, respectively.
This is impressive considering that the models for \tds have no access to labels for troll users.

Looking at individual features, for both T1 and T2, the embeddings from U2M outperform those from U2H and from BERT.
One possible reason is that the U2M graph is larger, and thus contains more information.
It is also possible that the social circle of a troll user is more indicative than the hashtags they used.
Finally, the textual content on Twitter is quite noisy, and thus the BERT embeddings perform slightly worse when used alone.

All our models with a single type of embedding easily outperform the model of \citet{kim2019tracking}. The difference is even larger when combining the embeddings, be it by concatenating the embedding vectors or by training separate models and then combining the posteriors of their predictions.

By concatenating the U2M and the U2H embeddings (U2H \concat U2M), we fully leverage the hashtags and the mention representations in the latent space, thus achieving accuracy of 88.7 for T1 and 78.0 for T2, which is slightly better than when training separate models and then averaging their posteriors (U2H \ensemble U2M): 88.3 for T1 and 77.9 for T2.
Adding BERT embeddings to the combination yields further improvements, and follows a similar trend, where feature concatenation works better, yielding 89.2 accuracy for T1 and 78.2 for T2 (compared to 89.0 accuracy for T1 and 78.0 for T2 for U2H \ensemble U2M \ensemble BERT).

Adding label propagation yields further improvements, both for LP1 and for LP2, with the latter being slightly superior: 89.6 vs. 89.3 accuracy for T1, and 78.5 vs. 78.3 for T2. 

Overall, our methodology achieves sizable improvements over previous work, reaching an accuracy of 89.6 vs. 84.0 of \citet{kim2019tracking} in the fully supervised case.
Moreover, it achieves 78.5 accuracy in the distant supervised case, which is only 11 points behind the result for \tsl, and is about 10 points above the majority class baseline.

\section{Discussion}
\label{sec:discuss}

\subsection{Ablation Study}

We performed different experiments with the hyper-parameters of the graph embeddings.
With smaller dimensionality (i.e.,~using 16 dimensions instead of 128), we noticed 2--3 points of absolute decrease in accuracy across the board.

Moreover, we found that using all of the data for learning the embeddings was better than focusing only on users that we target in this study, namely \emph{left}, \emph{right}, and \emph{news feed}, i.e.,~using the rest of the data adds additional context to the embedding space, and makes the target labels more contextually distinguishable.
Similarly, we observe 5--6 points of absolute drop in accuracy when training our embeddings on tweets by trolls labeled as \emph{left}, \emph{right}, and \emph{news feed}.

\subsection{Comparison to Full Supervision}

Next, we compared to the work of \citet{kim2019tracking}, who had a fully supervised learning scenario,
based on Tarde's Actor-Network Theory.
They paid more attention to the content of the tweet by applying a text-distance metric in order to capture the semantic distance between two sequences.
In contrast, we focus on critical elements of information that are salient in Twitter: \emph{hashtags} and \emph{user mentions}.
By building a connection between users, hashtags, and user mentions, we effectively filtered out the noise and we focused only on the most sensitive type of context, thus automatically capturing features from this network via graph embeddings.

\begin{table}[ht]
\begin{center}
    \begin{tabular}{l@{ }rr}
      \toprule
      \textbf{Method} & \textbf{Accuracy} &  \textbf{Macro F1}\\
        Baseline (majority) & 46.5 & 21.1\\
        BERT & 61.8 & 60.4\\
        U2H  & 61.6 & 60.0\\
        U2M  & 62.7 & 61.4\\
        U2H \ensemble U2M  & 63.5 & 61.8\\
        U2H \concat U2M  & 63.8 & 61.9\\
        U2H \ensemble U2M \ensemble BERT & 63.7 & 61.8\\
        U2H \concat U2M \concat BERT & 64.0 & 62.2\\       
      \bottomrule
    \end{tabular}
    \caption{\label{tab:mediastudy}Leveraging user embeddings to predict the bias of the media cited by troll users.}
\end{center}
\end{table}

\subsection{Reverse Classification: Media from Trolls}

Table~\ref{tab:mediastudy} shows an experiment in distant supervision for reverse classification, where we trained a model on the IRA dataset with the troll labels, and then we applied that model to the representation of the media in the MBFC dataset, where each medium is represented as the average of the embeddings of the users who cited that medium. We can see that we could improve over the baseline by 20 points absolute in terms of accuracy and by 41 in terms absolute in terms of macro-averaged F1.

We can see in Table~\ref{tab:mediastudy} that the relative ordering in terms or performance for the different models is consistent with that for the experiments in the previous section. This suggests that the relationship between trolls and media goes both ways, and thus we can use labels for media as a way to label users, and we can also use labels for troll users as a way to label media.

\section{Conclusion and Future Work}
\label{sec:conclusion}

We have proposed a novel approach to analyze the behavior patterns of political trolls according to their political leaning (\emph{left} vs. \emph{news feed} vs. \emph{right}) using features from social media, i.e., from Twitter. We experimented with two scenarios: (\emph{i})~supervised learning, where labels for trolls are provided, and (\emph{ii})~distant supervision, where such labels are not available, and we rely on more common labels for news outlets cited by the trolls. Technically, we leveraged the community structure and the text of the messages in the online social network of trolls represented as a graph, from which we extracted several types of representations, i.e.,~embeddings, for the trolls. Our experiments on the ``IRA Russian Troll'' dataset have shown improvements over the state-of-the-art in the supervised scenario, while providing a compelling case for the distant-supervision scenario, which has not been explored before.\footnote{Our data and code are available at \url{http://github.com/amatanasov/conll_political_trolls}}

In future work, we plan to apply our methodology to other political events such as \emph{Brexit} as well as to other election campaigns around the world, in connection to which large-scale troll campaigns have been revealed.
We further plan experiments with other graph embedding methods, and with other social media. Finally, the relationship between media bias and troll's political role that we have highlighted in this paper is extremely interesting.
We have shown how to use it to go from the media-space to the user-space and vice-versa, but so far we have just scratched the surface in terms of understanding of the process that generated these data and its possible applications. 

\section*{Acknowledgments}

This research is part of the Tanbih project,\footnote{\url{http://tanbih.qcri.org/}} which aims to limit the effect of ``fake news'', propaganda and media bias by making users aware of what they are reading. The project is developed in collaboration between the Qatar Computing Research Institute, HBKU and the MIT Computer Science and Artificial Intelligence Laboratory.

Gianmarco De Francisci Morales acknowledges support from Intesa Sanpaolo Innovation Center. The funder had no role in the study design, in the data collection and analysis, in the decision to publish, or in the preparation of the manuscript.

\bibliography{conll-2019}
\bibliographystyle{acl_natbib}

\end{document}